\documentclass[pdflatex,sn-mathphys-num]{sn-jnl}

\usepackage{graphicx}%
\usepackage{multirow}%
\usepackage{amsmath,amssymb,amsfonts}%
\usepackage{amsthm}%
\usepackage{mathrsfs}%
\usepackage[title]{appendix}%
\usepackage{xcolor}%
\usepackage{textcomp}%
\usepackage{manyfoot}%
\usepackage{booktabs}%
\usepackage{algorithm}%
\usepackage{algorithmicx}%
\usepackage{algpseudocode}%
\usepackage{listings}%

\usepackage{tikz}
\usetikzlibrary{arrows,shapes}
\usetikzlibrary{arrows.meta}
\usepackage{amsmath}
\usepackage{amssymb}
\usepackage{amsfonts}
\usepackage{xcolor}
\usepackage{graphicx}
\newcommand{\pr}{\mathbb{P}}

\usepackage{multirow}
\usepackage{float}
\usepackage{booktabs}
\usepackage{soul}
\usepackage{natbib}
\makeatletter
\providecommand{\@jmlrproceedings}{} 
\providecommand{\@jmlrvolume}{} 
\makeatother

\usepackage[utf8]{inputenc}
\usepackage{booktabs}%
\usepackage[utf8]{inputenc}
\usepackage{newunicodechar}
\newunicodechar{−}{-}

\numberwithin{equation}{section}

\theoremstyle{thmstyleone}%
\theoremstyle{thmstyletwo}%

\theoremstyle{thmstylethree}%

\begin{document}

\title[]{The Causal Case Against Algorithmic Discrimination: A Framework for Fairness Auditing in Law}



\author*[1]{\fnm{} \sur{Ruta Binkyte}}\email{ruta.binkyte@dfki.de}
\equalcont{These authors contributed equally to this work.}
\author[2]{\fnm{} \sur{Ljupcho Grozdanovski}}\email{grozdanovski@uliege.be}
\equalcont{These authors contributed equally to this work.}

\author[3]{\fnm{} \sur{Sami Zhioua}}\email{sami.zhioua@udst.edu.qa}

\affil*[1]{ \orgname{German Research Center for Arificial Intelligence (DFKI)}}
\affil*[2]{ \orgname{University of Liège}}
\affil*[3]{ \orgname{University of Doha for Science and Technology}}




\abstract{\begin{abstract} 
 TThe article addresses a central difficulty in contemporary regulation of artificial intelligence: legal standards for discrimination are grounded in causal reasoning, while most technical approaches to algorithmic fairness remain correlational. In discrimination law, liability typically depends on a counterfactual “but-for” inquiry, which determines whether an individual would have received a different outcome absent a protected characteristic. At the same time, computer science has developed sophisticated tools for causal inference in data analysis. Despite their conceptual alignment, these two domains have largely evolved independently. Our paper develops a unified framework that connects doctrinal legal causation with formal causal methods from statistics and machine learning, enabling causal evidence to inform the legal assessment of algorithmic discrimination.
The manuscript introduces two complementary forms of causal auditing. The first evaluates whether outcomes would differ if a sensitive attribute were changed, capturing direct and indirect discrimination. The second asks whether the outcome would have differed had the algorithm not been used at all, allowing courts and regulators to evaluate institutional responsibility for deploying automated decision systems. To operationalize this second inquiry, we propose a novel auditing metric comparing the total causal effect of a protected attribute on historical human decisions and on algorithmic predictions. Through mediation analysis and simulation experiments, we demonstrate how the method identifies hidden mechanisms of discrimination and remains applicable even for black-box models.
The paper also analyzes case law and emerging regulatory initiatives, including recent European AI governance proposals, to assess the legal admissibility of causal evidence and the evidentiary challenges created by algorithmic opacity. By integrating doctrinal legal analysis, causal inference methodology, and audit design, the article offers a practically actionable framework for attributing responsibility and evaluating fairness in automated decision-making.

\end{abstract}
}




\maketitle

\section{Introduction}\label{sec1}


As ML systems become increasingly influential, concerns about fairness, accountability, and transparency have taken center stage. Numerous international initiatives, including the EU AI Act, the OECD AI Principles, and national governance frameworks, have recognized fairness as a foundational requirement for trustworthy AI~\cite{EU_AI_Act_2021,OECD_AI_Principles_2019,G20_AI_Principles_2019,Singapore_AI_Governance_2020}. These policies emphasize that ML systems must avoid discriminatory behavior, particularly in decisions affecting individuals on the basis of protected attributes such as race, gender, or socioeconomic status.

The problem of discrimination is particularly pressing in the context of algorithmic decision-making. Despite widespread beliefs about algorithmic neutrality, prominent failures have shown that machine learning predictions can systematically disadvantage already marginalized groups. These instances reveal that biases embedded in training data, model design, or deployment environments can manifest as harmful outcomes for protected individuals.

Documented cases include biased healthcare risk predictions that undervalued the needs of Black patients~\cite{obermeyer2019dissecting}, lower facial recognition accuracy for women and minority groups~\cite{buolamwini2018gender}, gender discrimination in online job advertisements~\cite{lambrecht2019algorithmic}, and sexist biases embedded in language models~\cite{blodgett2020language}.

Algorithmic discrimination can be more harmful than traditional bias for two key reasons. First, the scale of algorithmic decisions means that millions of individuals can be affected in seconds. Second, the ML development cycle can reinforce these harms through feedback loops: biased outcomes are recorded, reused as training data, and further normalize discriminatory patterns.

To rigorously identify and mitigate these harms, causal reasoning has emerged as an indispensable tool. Unlike purely statistical fairness criteria, causal approaches allow for counterfactual evaluation: Would an individual have received a different outcome had they belonged to a different demographic group, all else being equal? This type of question is central not only to fairness in AI but also to legal standards for proving discrimination, which require causal evidence of harm.

Pearl’s structural causal model (SCM) framework provides a formal basis for such inquiries. It models causal relationships using directed acyclic graphs (DAGs) and allows analysts to estimate counterfactuals and intervention effects through tools like do-calculus. In contrast to correlation-based methods, SCMs enable reasoning about the mechanisms underlying observed disparities and about hypothetical alternatives.

Causal methods have gained traction in fairness auditing. Notable techniques include counterfactual fairness~\cite{kusner2017counterfactual}, mediation analysis~\cite{nabi2018fair}, and path-specific effects~\cite{chiappa2019path}, which make it possible to distinguish between justified and unjustified causal influence of sensitive attributes on outcomes. These tools help uncover biases that remain hidden under standard statistical checks and offer more nuanced ways to manage trade-offs between fairness and model performance~\cite{binkyte2025causality}.

This article contributes to this growing literature by developing a unified causal and legal framework for auditing algorithmic discrimination. We focus on two complementary forms of counterfactual reasoning grounded in the legal standard of \textit{but-for} causality. The first, which we term the “but-for sensitive attribute” framework, asks whether the outcome would have been different had the individual possessed a different value of a protected characteristic, such as race or gender. This test isolates direct or indirect discrimination by evaluating causal pathways involving sensitive attributes.

The second framework, introduced here as the “but-for algorithmic decision” test, shifts the focus from the individual to the technology: Would the decision have been different if the algorithm had not been used at all? This formulation captures institutional or system-level responsibility, especially in cases where algorithm deployment itself creates or amplifies existing biases. To operationalize this test, we propose a novel causal metric: the difference in total effect of the sensitive attribute on outcomes with and without the algorithm. This measure enables courts and regulators to assess whether the algorithm exacerbated, reduced, or replicated historical discrimination.

This article examines the challenges and opportunities involved in applying such causal frameworks to promote fairness in AI-driven decision-making from both legal and technical perspectives. We begin by reviewing foundational legal principles concerning causality and fairness, with a particular focus on procedural safeguards. We then analyze landmark discrimination cases, such as \textit{Cook v. HSBC}, \textit{Pickett}, \textit{Loomis}, and \textit{Ewert}, to demonstrate how current legal reasoning struggles to assess causation when facing algorithmic opacity.

Next, we introduce causal inference methodologies suited for fairness auditing, including mediation analysis, counterfactual estimation, and causal discovery. We show how these tools can support both types of “but-for” reasoning and clarify how they relate to emerging legal standards, such as those proposed in the EU AI Act and AI Liability Directive.


\subsection*{Main Contributions}

This paper makes several key contributions to the study of auditing fairness in AI through a causal inference perspective. Specifically, we:

\begin{itemize}
\item \textbf{Establish the connection between causal fairness analysis and legal frameworks.} We analyze how causal methods align with judicial practices, particularly within the European legal framework, and discuss the admissibility and role of causal evidence in proving algorithmic discrimination in court. 

\item \textbf{Define "but-for" algorithmic decision legal framework.} Based on recent court practice, we define how legal causal reasoning is evolving—from focusing on 'but-for' sensitive attributes to emphasizing 'but-for' algorithmic decisions in causal discrimination cases.

\item \textbf{Introduce causal auditing metric compatible with the "\textit{but for}" algorithmic decision causal framework.} We propose a clear and legally interpretable framework that compares the total effect of a sensitive attribute on historical outcomes versus algorithmic predictions. \\

\end{itemize}

In addition to our theoretical framework, we present illustrative experiments using synthetic data to demonstrate how causal discrimination can be decomposed and attributed to algorithmic decision-making. These examples clarify the practical computation of our proposed metric and highlight subtleties in interpreting fairness through a legal lens. 


\subsection*{Paper Structure}
This paper is structured as follows. In \autoref{sec:background} we introduce main concepts of statistical causality. In \autoref{sec:legal} we bridge legal definition of causality with the statistical causal framework. In~\autoref{sec:mediation} we provide a case study for fine-grained causal discrimination analysis. In~\autoref{sec:challenges} we discuss challenges in legally proving causality in algorithmic discrimination cases. In~\autoref{sec:butfor} we introduce a shift from "\textit{but for}" sensitive attribute to "\textit{but for}" algorithmic decision framework in the recent court practices. In~\autoref{sec:metric} we define a causal metric for causal discrimination auditing for the black-box algorithms. In~\autoref{sec:evidence} we discuss the legal aspects of the access to evidence necessary for the algorithmic auditing. In~\autoref{sec:constraints} we discuss practical constraints of applying the proposed metric and offer possible solutions. Finally, we conclude in \autoref{sec:conclusion}.

\section{Related Work}
Over the last decade, causal fairness has emerged as an important approach in the quest for equitable AI. One of the most influential works in counterfactual fairness, introduced by Kusner et al~\cite{kusner2017counterfactual}. They formalized the idea that a decision is fair “if it is the same in the actual world and in a counterfactual world where the individual belonged to a different demographic group”. Building on this, subsequent research refined causal fairness notions by distinguishing fair and unfair pathways of influence. Building on this, Nabi et al.\cite{nabi2018fair} framed discrimination as causal effects along impermissible paths. Chiappa et al.\cite{chiappa2019path} further developed methods to quantify these path-specific effects. Zhang et al~\cite{zhang2018fairness} further contributed a causal explanation formula to decompose observed outcome disparities into portions attributable to different causal factors, enabling one to explain and quantify unfairness in decision-making. Loftus et al~\cite{loftus2018causal} argued for the importance of causal reasoning when designing fair algorithms, pointing out that purely correlation-based fairness criteria may be misleading if the data is biased by historical discrimination. Plevcko et al~\cite{plevcko2024causal} enumerates which causal assumptions are needed to test for disparate impact versus disparate treatment and connect causal fairness to the legal disparate treatment framework.

Legal scholarship has increasingly recognized the importance of causality in assessing algorithmic fairness. Carey et al~\cite{carey2022fieldguide} discuss the relationship between causality-based fairness notions to legal principles, noting that many fairness criteria in machine learning echo concepts from discrimination law when interpreted causally. However, they do not link causal fairness in ML and court practice in the context of AI auditing. Xiang~\cite{xiang2021reconciling} analyzes the compatibility of technical approaches to algorithmic bias with U.S. anti-discrimination law, highlighting the potential of causal inference to reconcile algorithmic fairness techniques with legal standards. Binkyte et al. examine the challenges and opportunities involved in applying principles of causality to ensure fairness in AI-driven decision-making from both legal and technical perspectives~\cite{binkyte2022need}. Weerts et al~\cite{weerts2023algorithmic} analyze seminal examples of algorithmic unfairness through the lens of EU non-discrimination law, drawing parallels with EU case law and highlighting the normative underpinnings of fairness metrics and technical interventions. These legal perspectives underscore the necessity of integrating causal analysis into algorithmic auditing to meet evidentiary standards and ensure compliance with anti-discrimination laws.

\section{Background on Statistical Causality}~\label{sec:background}

 Here we provide a technical preliminaries for causal framework and causal fairness notions~\citep{pearl2009causality}.

\subsection{Causal Graph}
\label{sec:caus}

\begin{figure}[H]
\begin{minipage}{0.32\linewidth}
\centering
		{\includegraphics[scale=0.6]{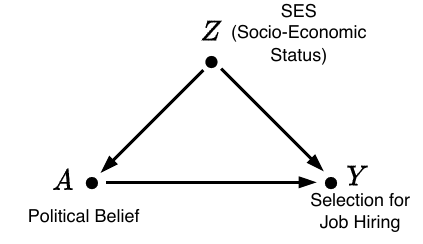}}
	\caption{Confounder structure.}
	\label{fig:simpleConfJobHiring}
\end{minipage}
\hspace{0.3cm}
\begin{minipage}{0.3\linewidth}
\centering
  \includegraphics[scale=0.6]{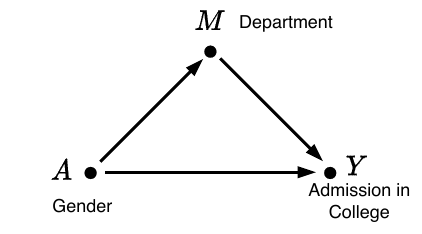}
	\caption{Mediator structure.}
	\label{fig:simpleMediatorAdmission}
 \end{minipage}
 \hspace{0.3cm}
\begin{minipage}{0.3\linewidth}
\centering
  \includegraphics[scale=0.6]{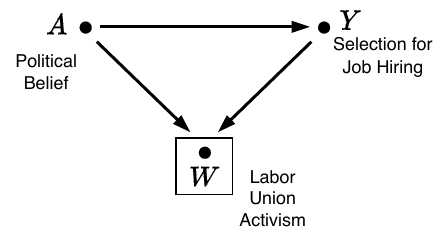}
	\caption{Collider structure.}
		\label{fig:simpleColliderJobHiring}
\end{minipage}
\end{figure}

A causal graph, denoted as \(\mathcal{G} = (\mathbf{V}, \mathcal{E})\), is a Directed Acyclic Graph (DAG) consisting of a set of variables or nodes \(\mathbf{V}\) and edges \(\mathcal{E}\). Each edge \(X \to Y\) signifies a causal relationship, meaning changes in \(X\) directly influence \(Y\). Importantly, altering \(X\) impacts \(Y\), but modifying \(Y\) does not affect \(X\). Figures~\ref{fig:simpleConfJobHiring}-\ref{fig:simpleColliderJobHiring}  show three basic DAG structures - confounder, mediator, and collider, that, if present, can result in statistical anomalies, making statistical fairness metrics unreliable.

\noindent

Detailed analysis of the effect of confounder, collider, and mediator structures on fairness measurement can be found in the Appendix~\ref{app:causalstructures}.







\subsection{Causal Fairness Notions}

Causal fairness aims to ensure that sensitive attributes, such as race or gender, do not unfairly influence outcomes. Below, we describe key causal fairness notions and their formal definitions.

\textbf{Total Effect (TE).}
Total Effect (TE)~\citep{pearl_causality_2009} is a causal fairness notion that quantifies the overall effect of a sensitive attribute \(A\) on an outcome \(Y\). Formally, TE is defined as:
\begin{equation}
\label{eq:TE}
TE = \pr(Y = y \mid do(A = a_1)) - \pr(Y = y \mid do(A = a_0)),
\end{equation}
where \(do(A=a)\) denotes an intervention that sets \(A\) to \(a\). TE measures the causal impact of changing \(A\) from \(a_0\) to \(a_1\) on \(Y\) across all causal paths connecting \(A\) to \(Y\). Using a more concise notation, TE (Equation~\ref{eq:TE}) can be simply expressed as 
\begin{equation}
\label{eq:TE2}
TE = \pr(y_{a_1}) - \pr(y_{a_0}).
\end{equation}

\textbf{Mediation Analysis (NDE and NIE).}
Mediation analysis decomposes the causal effect of \(A\) on \(Y\) into direct and indirect effects. This is essential for identifying the pathways through which \(A\) influences \(Y\).

\textit{Natural Direct Effect (NDE})~\citep{pearl01direct}:  
The NDE quantifies the direct effect of \(A\) on \(Y\), bypassing any mediators. For a binary variable \(A\) with values \(a_0\) and \(a_1\), the NDE is:
\begin{equation}
\label{eq:NDE}
NDE = \pr(y_{a_1, \mathbf{M}_{a_0}}) - \ \pr(y_{a_0, \mathbf{M}_{a_0}})
\end{equation}
where \(\mathbf{M}\) represents the set of mediator variables, and \(\pr(y_{a_1, \mathbf{M}_{a_0}})\) is the probability of \(Y=y\) if \(A\) is set to \(a_1\) while the mediators are set to values they would take under \(M=a_0\).

\textit{Natural Indirect Effect (NIE ($M_i$))}~\citep{pearl01direct}:  
The NIE($M_i$) captures the influence of \(A\) on \(Y\) through the mediation of $M_i$. In case of two binary mediators $M_i$ and $M_j$ it is given by:


\begin{equation}
\label{eq:NIE}
NIE = \pr(y_{a_0, \mathbf{M_i}_{a_1},\mathbf{M_j}_{a_0}}) - \ \pr(y_{a_0, \mathbf{M_i}_{a_0},\mathbf{M_j}_{a_0}})
\end{equation}

where $\pr(y_{a_0, \mathbf{M_i}_{a_1},\mathbf{M_j}_{a_0}})$ denotes the probability of \(Y=y\) with the sensitive attribute \(A\) set to \(a_0\), the mediator \(M_i\) takes the value it would take under \(A = a_1\), and the other mediator \(M_j\) is held at the value it would take under \(A = a_0\). This expression quantifies the \emph{natural indirect effect} of \(A\) on \(Y\) through \(M_i\), while holding all other mediators fixed at their baseline (untreated) levels.

\textbf{Path-specific effect (PSE).} Path-specific effect~\cite{pearl2009causality,chiappa2019path,wu2019pc} is a more nuanced measure that characterizes the causal effect in terms of specific paths.

Given a path set $\pi$, the $\pi$-specific effect is defined as: 
\begin{equation}
\label{eq:PSE}
\mbox {PSE}^{\pi} = \pr(y_{a_1 |_\pi, a_0 |_{\overline{\pi}}}) - \pr(y_{a_0})
\end{equation}
where $\pr(y_{a_1 |_\pi, a_0 |_{\overline{\pi}}})$ is the probability of $Y=y$ in the counterfactual situation where the effect of $A$ on $Y$ with the intervention ($a_1$) is transmitted along $\pi$, while the effect of $A$ on $Y$ without the intervention ($a_0$) is transmitted along paths not in $\pi$ (denoted by: $\overline{\pi}$).



\section{From Legal to Statistical Causality}~\label{sec:legal}
\label{sec:legalNew}

As a preliminary clarification, in legal contexts the term \textit{causal fairness} does not refer to fairness in general, but to the procedural conditions under which claims of fairness or unfairness are evaluated through causal reasoning. This distinction matters because legal systems typically require that such claims be supported by causal evidence, establishing that a harm or benefit can be attributed to a specific factor. With this in mind, the discussion that follows focuses on two critical and interrelated components of causal fairness in law: \textit{evidence} and \textit{procedural fairness}.%
\footnote{In the EU, the Independent High-Level Expert Group on AI, established by the European Commission, defines procedural fairness as the ability to contest and seek effective redress against decisions made by AI systems and their human operators. See HLEG, \textit{Ethics Guidelines for Trustworthy AI}, available at \url{https://digital-strategy.ec.europa.eu/en/library/ethics-guidelines-trustworthy-ai}}

\begin{table}[h!]
\centering
\begin{tabular}{|p{2.5cm}|p{5.1cm}|p{5.1cm}|}
\hline
\textbf{Aspect} & \textbf{Causality in Computer Science} & \textbf{Causality in Law} \\
\hline
\textbf{Purpose} & Discover and model causal relationships to support predictions, explanations, interventions, and fairness evaluation & Establish legal responsibility or liability by proving that an action or factor caused harm \\
\hline
\textbf{Typical Frameworks} & Structural Causal Models (Pearl), Potential Outcomes (Rubin), do-calculus, counterfactual estimation & “But-for” test, proximate cause, legal standards of proof (e.g., preponderance of evidence) \\
\hline
\textbf{Evidence Base} & Observational or experimental data analyzed with causal graphs, statistical methods & Testimonies, documentation, statistical or expert evidence, procedural rules of admissibility \\
\hline
\textbf{Key Question} & Does changing \( X \) (e.g., gender) affect \( Y \) (e.g., prediction or outcome), holding other factors constant? & Would the harm have occurred \textit{but for} the presence of factor \( X \)? Can a reasonable chain of causation be established? \\
\hline
\textbf{Counterfactuals} & Formalized using causal models and potential outcomes; estimated from data with assumptions & Hypothetical reconstructions based on logic, reasoning, expert judgment, and procedural standards \\
\hline
\textbf{Assumptions Required} & Explicit assumptions about confounders, causal graphs, no unmeasured confounding, SUTVA & Legal presumptions (e.g., foreseeability, duty of care), often context-dependent and debated \\
\hline
\textbf{Decision Goal} & Scientific inference, prediction improvement, fairness auditing, accountability design & Legal judgment on liability, remedy, or responsibility \\
\hline
\end{tabular}
\caption{Contrasting Causality in Computer Science and Law}
\label{tab:causality_comparison}
\end{table}

From a legal perspective, causality is fundamentally a question of fact. Establishing causation requires not only evidentiary procedures defined by law, but also appropriate reasoning models capable of distinguishing true causal relationships from mere statistical associations or correlations, such as observed patterns between candidate causes and a given harm~\cite{haack2014evidence}.

Importantly, in legal adjudication, causality is not pursued for its own sake but in service of fairness, typically as a basis for determining whether compensation or some other remedy is warranted. Legal systems grounded in the rule of law%
\footnote{In the EU, the concept of the rule of law encompasses the following principles: legality, legal certainty, prohibition of arbitrariness by executive powers, independent and impartial courts, effective judicial review (including respect for fundamental rights), and equality before the law. See Communication from the Commission to the European Parliament and the Council, ‘A New EU Framework to Strengthen the Rule of Law,’ COM(2014) 158 final, p. 4.} 
place a strong emphasis on procedural fairness - the normative principle that only fair procedures can lead to just outcomes.

This relationship between process and outcome is institutionalized in modern evidentiary and judicial systems, including those in EU law. Here, procedural fairness is reflected in safeguards such as the right to a fair trial, which encompasses not only equal access to justice, but also equal access to relevant evidence. The underlying idea is that no party should be advantaged or disadvantaged in their ability to present the facts necessary to make their case before a court~\cite{solum2004procedural}. This principle of \textit{procedural parity} aims to ensure that each side has a fair opportunity to be heard.

In short, achieving causal fairness under the law depends on both the availability of credible causal evidence and the conditions under which that evidence is obtained and presented. What matters is not just whether causality can be established, but whether it can be established fairly~\cite{rescher1980plausible}.

To support this effort, law can draw on formal tools developed in the field of statistical causality. Although the underlying goals and standards differ, both legal and computational approaches aim to determine whether one factor can be said to have caused another in a manner that justifies responsibility or intervention. A comparison of how causality is treated in both fields is presented in Table~\ref{tab:causality_comparison}.

Two leading frameworks for causal inference have emerged from the statistical literature: the potential outcomes framework, introduced by Rubin~\cite{rubin2005causal}, and the structural causal model (SCM) framework developed by Pearl~\cite{pearl2009causality}. Rubin's approach emphasizes counterfactual reasoning grounded in randomized experimentation, while Pearl's framework uses graphical models and do-calculus to analyze causal relationships in complex systems.

Both frameworks help clarify whether an observed association, such as between a sensitive attribute (e.g., gender or race) and an outcome (e.g., hiring or credit approval), reflects a true causal influence or is merely a statistical artifact. This distinction is crucial in fairness analysis, where surface-level correlations may be misleading. A classic example is the Berkeley admissions case~\cite{berkeley75}, where initial statistics suggested gender bias in graduate admissions, but deeper causal analysis revealed that women disproportionately applied to departments with lower acceptance rates, highlighting confounding rather than actual discrimination.

Causal reasoning is especially vital in machine learning (ML) fairness, where decisions are often based on historical, observational data. Statistical fairness metrics, such as demographic parity or equalized odds, fail to uncover true discrimination when historical inequities are embedded in the data. In contrast, causal fairness methods allow analysts to break down the total effect of a sensitive attribute into direct and indirect (mediated) pathways. This helps distinguish between justified disparities (e.g., arising from legitimate qualifications) and unjustified ones (e.g., stemming from redlining practices)~\cite{makhlouf2024causality, binkyte2025causality}.

These tools are increasingly applied to audit datasets used in algorithmic systems. Measures like the total effect (TE) and natural direct effect (NDE) provide insight into whether a protected attribute has had a causal impact on decisions, either directly or through impermissible proxies~\cite{chiappa2019path}. Such insights are critical in legal contexts, where establishing the causal origin of harm is essential to assigning responsibility and enabling remediation.

In the next section, we apply these ideas through a case study, demonstrating how causal fairness analysis can help disentangle causal from non-causal relationships and distinguish between discriminatory and justifiable effects linked to sensitive attributes.

\section{Causal Discrimination Analysis}~\label{sec:mediation}
\label{sec:medAnalysis}

\begin{figure}
    \centering
    \includegraphics[width=\linewidth]{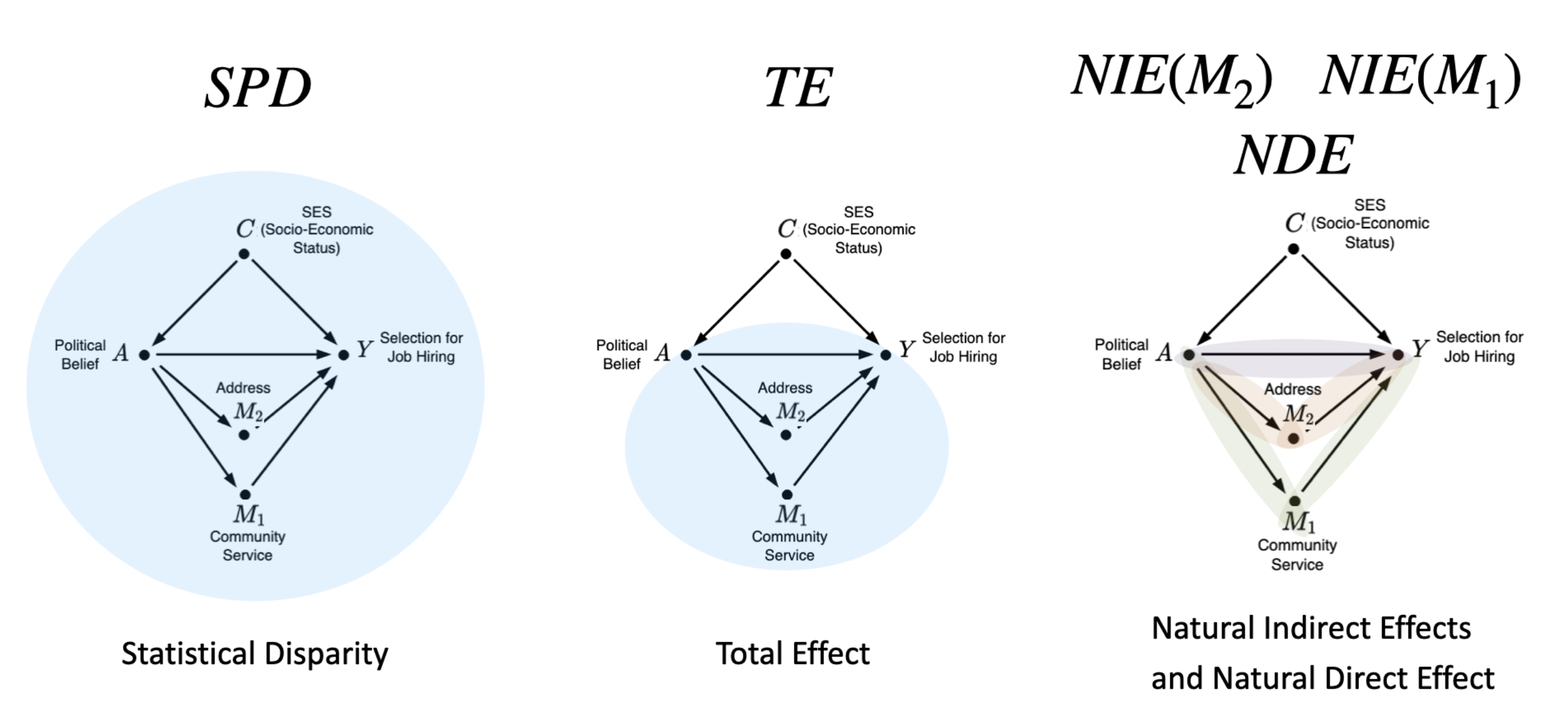}
    \caption{The causal graph describes a situation with direct path between the sensitive attribute $A$ and the outcome $Y$, confounding path between the $A$ and $Y$ through $C$ and two mediated paths through $M_1$ and $M_2$. The shaded areas represent the scope of non-causal ($SPD$) and causal ($TE$, $NDE$, $NIE (M_1)$, and $NIE(M_2)$) fairness metrics. }
    \label{fig:mediationAnalysis}
\end{figure}

In presence of multiple causal and non-causal paths linking sensitive attributes and the outcome, it can be difficult to distinguish discrimination from justifiable effect or spurious correlations. Mediation analysis aims to distinguish the different causal pathways through which discrimination propagates and quantify the proportion of discrimination conveyed through each path.  

Consider a variation of the job hiring example, illustrated in Figure~\ref{fig:mediationAnalysis}, which includes two mediator variables:  community service participation ($M_1$) and address ($M_2$). In this scenario, there are four distinct causal paths from the sensitive variable $A$ (political belief) to the outcome variable $Y$ (job hiring):

\begin{itemize}
    \item $A \leftarrow C \rightarrow Y$: confounding path
    \item $A \rightarrow Y$: direct path
    \item $A \rightarrow M_1 \rightarrow Y$: indirect path through $M_1$
    \item $A \rightarrow M_2\rightarrow Y$: indirect path through $M_2$.
\end{itemize}
The first confounding path is non-causal, meaning that any effect transmitted through it should not be considered when estimating discrimination. 
This spurious effect arises from the way data is generated or collected and, therefore, should not be counted as actual discrimination.  

The direct path exists whenever there is an edge between $A$ and $Y$. The effect through $A \rightarrow Y$ is always discriminatory. Namely, it can never be justified or considered acceptable discrimination.  

The two remaining paths are indirect paths passing through mediator variables. Whether discrimination along an indirect path is justifiable depends on the nature of the mediator variable. For instance, in the job hiring example shown in Figure~\ref{fig:mediationAnalysis}, $M_2$ (home address) acts as a mediator because, on one hand, political inclination may influence where a candidate lives, and on the other hand, hiring decisions may be influenced by a candidate's address. Similarly, $M_1$ (community service) is a mediator because a candidate's political beliefs may impact their level of involvement in community service, while an employer may consider community service records as an indicator of suitability for a given position.  

Discrimination along the path $A \rightarrow M_1 \rightarrow Y$ can be considered justifiable, as an employer may argue that disparities between candidates with different political beliefs arise from differences in their community service records. However, discrimination through the path $A \rightarrow M_2 \rightarrow Y$ is typically not acceptable, as an employer cannot justify hiring decisions based on candidates' addresses. In this context, $M_2$ is referred to as a \textit{proxy variable}, whereas $M_1$ is called an \textit{explaining variable}.\footnote{If a single causal path consists of two or more mediator variables, the presence of at least one explaining variable among the mediators makes discrimination along that path justifiable and, therefore, acceptable.}

Causality, through the concepts of intervention and counterfactual reasoning, provides the necessary tools to distinguish between different types of discrimination based on their causal pathways. By intervening on $A$, all paths that include an incoming edge to $A$, such as confounding paths between $A$ and $Y$, are blocked. Discrimination transmitted through causal paths is quantified by the \textit{total effect} ($TE$) (~\autoref{eq:TE}). 
In the causal graph depicted in Figure~\ref{fig:mediationAnalysis}, the $TE$ expression captures discrimination along all causal paths except for the non-causal confounding path $A \leftarrow Z \rightarrow Y$.  



To distinguish between direct and indirect discrimination, two key expressions can be used: the \textit{natural direct effect} ($NDE$) (\autoref{eq:NDE}) and the \textit{natural indirect effect} ($NIE$) (\autoref{eq:NIE})~\citep{pearl01direct}. 
Intuitively, $NDE$ is counterfactual quantity because it corresponds to a candidate who is conservative ($A=1$) along the direct path $A\rightarrow Y$ but liberal ($A=0$) along all indirect paths. This isolates the direct effect of $A$ on $Y$ while holding indirect pathways constant.  



Finally, discrimination conveyed through specific indirect paths can be identified using the \textit{Natural Indirect Effect through mediator $M_i$} ($NIE (M_i)$) (\autoref{eq:NIE}) ~\citep{pearl01direct,chiappa2019path}. For example, $NIE (M_2)$) quantifies the extent to which a sensitive attribute influences a decision through the mediator "address" compared to the mediator 'community service.' This has direct implications for fairness assessment. 



\paragraph{\textit{Empirical Illustration of Mediation Analysis.}} To illustrate a mediation structure typical of algorithmic decision-making, we construct a synthetic hiring scenario~\footnote{This example is based on simulated data to illustrate causal auditing methods and is not drawn from real-world case studies.}. Each individual is described by a sensitive attribute $A$ (e.g., political affiliation), a socioeconomic status variable $C$, and two mediators: community service hours ($M_1$) and urban/rural residency ($M_2$). The hiring decision $Y$ is modeled using logistic regression, influenced by these variables. The synthetic data generation process is detailed in Appendix~\ref{app:dgp}.

The simulation embeds both direct discrimination (via a negative coefficient on $A$) and indirect effects through $M_1$ and $M_2$. For instance, liberals ($A = 1$) tend to perform more community service, which positively affects hiring. However, community service is itself influenced by SES ($C$), which negatively impacts hiring, capturing the real-world complexity where protected attributes correlate with seemingly legitimate factors in decision-making.

We first compare observed hiring rates: liberals are hired less often than conservatives (28.9\% vs. 37.3\%), producing a raw disparity of $-0.084$. However, this statistical parity measure conflates multiple causal mechanisms and ignores the data-generating structure.

Next, we apply mediation analysis to estimate causal effects. The total effect (TE) of being liberal on hiring is positive (+0.0272), despite a negative direct effect (−0.1173). This is explained by a strong positive indirect effect via community service ($M_1$, +0.1349), and a smaller positive effect via residency ($M_2$, +0.0271). The total effect thus reflects a balance between direct disadvantage and compensating mediators.

From a fairness perspective, this decomposition highlights the limits of relying solely on raw disparities or direct effect estimates. Although the liberal label directly reduces hiring chances, liberal candidates perform more community service, which offsets some of the bias. Moreover, SES, which is omitted from naive comparisons, plays a major confounding role: liberals are more likely to be lower SES, which strongly depresses their hiring prospects.

This example underscores the value of causal modeling in fairness analysis. It shows how structural inequalities, mediated through legitimate or semi-legitimate variables, can obscure the sources of observed disparities. Comparing group outcomes alone risks attributing inequality to discrimination when deeper causes lie in social structure.

Ultimately, while liberals are hired less often in the observed data, this disparity is largely explained by socioeconomic disadvantage, instead of the political label. A direct bias against liberals is still present, but whether community service should be seen as a fair mitigating factor remains a normative question. The details of the implementation can be found in~\autoref{app:exp_design}, ~\autoref{app:dgp} and~\autoref{app:mediation_est}.

\section{Challenges in Auditing Fairness in Algorithmic Discrimination Cases}\label{sec:challenges}

\textbf{\textit{Proof of Causality in Legal Disputes.}} The challenge of proving causation in algorithmic discrimination cases has significantly strained long-standing legal doctrines. From a procedural fairness perspective, a core difficulty stems from the opacity - whether partial or complete - of AI systems, making them causally non-deterministic and hindering any reasonable assessment and foresight of their ‘causal power’~\cite{mackie1965}. Indeed, standard liability doctrines rely on a Humean vision of reality which posits that any real-world phenomenon can be explained through causal networks that operate under laws of causal necessity (or ‘regularity’)~\cite{hume1888}.

Although the historic stance that reality is causally deterministic has been much debated since Hume~\cite{kistler2006}, it continues to, nevertheless, represent the theoretical cornerstone of modern liability law and doctrine, underpinning two general (and generalized) procedural assumptions. First, the assumption that any real-world phenomenon (like the suffering of harm) can be causally explained through the search for, and proof of a ‘necessary and sufficient cause’~\cite{harthonore1985, white1984}. This generates the second and related assumption that the evidence and casual explanations of harm suffered are usually within the litigants’ reach. As stressed by Cheng et al., in evidence law ‘ought’ implies ‘can’~\cite{chengpardo2015} meaning that the evidence the law requires for the proof of specific facts is generally assumed to be feasible for litigants. 

AI has deeply challenged both assumptions i.e. the knowability of harm and causation, on the one hand and their provability, on the other hand. The topical literature on the so-called algorithmic discrimination~\cite{xenidis2021,xenidis2020,xenidis2023} tends to flag, as root cause of these challenges, the fact that AI systems discriminate by proxy, connecting variables which, under a veneer of neutrality, in fact lead to discriminatory outcomes.

An example in that connection is \textit{Cook v. HSBC} North America,~\cite{cookvhsbc2014} a credit-scoring case in which the system factored in applicants’ places of residence, disproportionately benefiting predominantly white neighborhoods while disadvantaging ethnic minorities. The combination of subtly discriminatory inputs - such as zip code and ethnicity~\cite{chen2024} - with limited transparency (ergo knowability) about how input variables were weighted, seriously undermined fundamental procedural safeguards, particularly the rights to access evidence and courts.

In response to these challenges, regulators and courts in and outside of Europe have begun addressing two pivotal questions: What evidence must litigants access to effectively prove causation? Once that evidence is identified, how should legal procedures be (re)structured to guarantee such access? These questions are central to adapting legal frameworks for algorithmic accountability while upholding procedural fairness.

With respect to the first question, the emerging - though still unsettled - case law on AI liability reveals a notable trend. While numerous cases could be cited, we focus here on three that exemplify a shifting approach to proving causation in AI-related disputes: Pickett~\cite{pickett2021}, which involves a DNA matching system (TrueAllele) used to identify the perpetrator of a crime; Loomis~\cite{loomis2016}, which concerns the use of COMPAS, a recidivism-prediction tool; and Ewert~\cite{ewert2018}, which addresses recidivism prediction in Canadian correctional services.

In all three cases, the plaintiffs alleged that the automated systems produced inaccurate outcomes due to unfair bias - specifically, gender bias in \textit{Pickett and Loomis}, and ethnic bias in Ewert. To expose these biases, the plaintiffs requested access to the systems’ inner workings, asking that the algorithms be reverse-engineered so that they could uncover the causal links between inputs and outcomes in view of assessing the accuracy of those outcomes with regard to the claimants’ individual cases. However, this proved practically infeasible. In Pickett, in particular, independent experts testified that reverse-engineering the TrueAllele system would take up to 8.5 years to complete~\cite{pickett2021}.

The emerging field of mechanistic interpretability offers tools for inspecting internal representations of models, including large language models, without requiring full reverse-engineering~\citep{bereska2024mechanistic}. However, establishing a robust causal link between these internal structures and external decision-making remains a major open problem~\citep{sharkey2025open,binkyte2025causality}.
Given the impracticality of full system transparency, the courts in \textit{Pickett and Loomis} adopted a fallback strategy: reliance on general expert testimony. In the absence of direct technical evidence, they turned to pre-existing expert evaluations of the systems’ overall reliability. If experts broadly agreed that a system - such as TrueAllele or COMPAS - performed accurately and fairly in general, the courts were willing to presume that the specific decisions under challenge were also unbiased~\footnote{It was argued that this was a shift from ‘what experts prove’ (essentially relating to the availability and trustworthiness of expertise in instances involving individual AI output) to ‘what experts say’ (relating to general opinions about the aptitude for accuracy of specific AI systems~\cite{grozdanovski2024, grozdanovski2025}.}.
This approach highlights the evidentiary gap at the heart of the budding AI litigation: when algorithmic opacity prevents direct examination of an output’s accuracy, fairness and causal explanation, courts may rely on indirect assurances, which shifts the burden from empirical validation to institutional trust. As we explore in the next section, addressing this evidentiary gap requires not only legal innovation but also methodological support - especially from tools rooted in causal inference. These tools offer new ways to approximate, interpret, and evaluate algorithmic decisions in light of fairness norms, even when full transparency is unattainable.

\section{From \textit{\textit{"but for"}} Sensitive Attribute to \textit{\textit{"but for"}} Algorithmic Decision}\label{sec:butfor}

The difficulty of proving causation through reliable evidence has led to a notable reinterpretation of the so-called \textit{but-for test}~\cite{moore2009}. Traditionally, this test relies on counterfactual reasoning to assess whether a harm would have occurred in the absence of a specific cause. For example, in \textit{Cook v. HSBC} (a credit scoring case), a standard application of the but-for test would ask whether the same loan applicants would have been approved had the system not taken their place of residence into account.

However, the cases discussed in the previous section - particularly \textit{Pickett and Loomis} - demonstrate a shift in how this test is applied in AI-related disputes. In conventional (non-AI) contexts, the but-for test addresses a factual counterfactual: would the outcome have changed if a particular feature (such as address, gender, or age) had not been considered? By contrast, in disputes involving algorithmic systems, the question is reframed: \textit{would a human decision, based on AI-generated output, have been the same if the AI system had not been used at all?}

This shift reflects a broader movement away from individual-level assessments of discrimination based on protected characteristics toward systemic evaluations of institutional responsibility for deploying potentially biased technologies. A structured comparison of these two applications of the \textit{but-for} test - highlighting their respective causal queries, legal significance, and evidentiary requirements - is provided in~\autoref{tab:butfor_comparison}.Within this evolving context, tools from statistical causality offer valuable means to detect discrimination in historical data used for hiring or lending decisions. If a significant association between a sensitive attribute and an outcome is identified, causality analysis can support the inference that the same biased result might have occurred even without the algorithm - implicating broader company practices rather than the algorithm itself. In this way, causality methods help distinguish between spurious correlations, explainable disparities, and genuine discrimination.

Conversely, if the data used to train an AI system encodes structural or historical biases, compliance with AI design and audit standards requires further scrutiny. Causal tools can then help determine whether any disparities found in outputs arise from unjustified use of sensitive information or proxies.

Moreover, causality frameworks provide precise mathematical expressions for otherwise abstract counterfactual questions~\citep{shpitser07-counterfactuals,shpitser08}, enabling formal verification of the but-for test in algorithmic settings.

This brings us back to the second foundational question raised above: should procedural systems of evidence explicitly include a right to request access or disclosure of evidence related to an AI system’s design and functioning? Addressing this question is crucial for aligning evolving legal standards with the practical requirements of causal proof in the age of algorithmic decision-making.

\section{Causal Black Box Auditing Based on \textit{but-for} Algorithmic Decision Legal Framework}
\label{sec:metric}

\textbf{\textit{Causal Auditing of Algorithmic Decisions.}} 
Building on the \textit{but-for} test used in legal reasoning, we propose a causal auditing methodology that distinguishes the effects of an algorithm from those of historical decision-making. This is achieved by comparing the total causal effect of a protected attribute \( A \) on two outcomes: the original human decision \( Y \) and the algorithmic prediction \( \hat{Y} \). While both reflect judgments about the same task (e.g., hiring), we treat them as distinct endpoints, enabling an audit of the algorithm’s specific contribution to disparity.

We define the total effect (TE) of \( A \) on each outcome as:
\begin{align}
TE_Y &= P(Y \mid do(A = 1)) - P(Y \mid do(A = 0)) \label{eq:te_y} \\ \nonumber \\
TE_{\hat{Y}} &= P(\hat{Y} \mid do(A = 1)) - P(\hat{Y} \mid do(A = 0)) \label{eq:te_yhat}
\end{align}

These effects are estimated using standard tools from causal inference, requiring identification of relevant covariates and the use of techniques such as counterfactual simulation or causal graphs. Importantly, this framework does not require access to the algorithm's internal logic and is compatible with black-box models.

To support legal interpretation and regulatory review, we categorize algorithmic behavior based on how \( TE_{\hat{Y}} \) compares to \( TE_Y \). The following taxonomy provides a principled way to assess whether the algorithm introduces, mitigates, or reverses disparities:

\begin{enumerate}
    \item \textbf{Match (Replication):} \\
    \( TE_{\hat{Y}} = TE_Y \) \\
    The algorithm reproduces the disparity observed in historical decisions.

    \item \textbf{Amplification:} \\
    \( |TE_{\hat{Y}}| > |TE_Y| \) and \( \operatorname{sign}(TE_{\hat{Y}}) = \operatorname{sign}(TE_Y) \) \\
    The algorithm increases the magnitude of the disparity in the same direction.

    \item \textbf{Mitigation:} \\
    \( 0 < |TE_{\hat{Y}}| < |TE_Y| \) and \( \operatorname{sign}(TE_{\hat{Y}}) = \operatorname{sign}(TE_Y) \) \\
    The algorithm reduces but does not eliminate the disparity.

    \item \textbf{Reversal:} \\
    \( \operatorname{sign}(TE_{\hat{Y}}) \ne \operatorname{sign}(TE_Y) \) and \( TE_{\hat{Y}} \ne 0 \) \\
    The algorithm reverses the direction of the disparity.

    \item \textbf{Neutralization:} \\
    \( TE_Y \ne 0 \), but \( TE_{\hat{Y}} = 0 \) \\
    The algorithm neutralizes the disparity, erasing evidence of it in its predictions.

    \item \textbf{Spurious Disparity:} \\
    \( TE_Y = 0 \), but \( TE_{\hat{Y}} \ne 0 \) \\
    The algorithm introduces disparity not present in the underlying human decisions.

    \item \textbf{No Disparity (Baseline Fairness):} \\
    \( TE_Y = 0 \) and \( TE_{\hat{Y}} = 0 \) \\
    Neither historical decisions nor algorithmic predictions show disparity with respect to \( A \).
\end{enumerate}

This classification is interpretable, audit-friendly, and legally actionable. It avoids relying on abstract or opaque metrics, instead emphasizing direct comparisons grounded in the counterfactual legal framework. In doing so, it supports evidence-based evaluations of algorithmic fairness in judicial and policy contexts. For the rest of the paper, we refer to such framework as $\Delta_{\text{alg}}$.

From a legal standpoint, the $\Delta_{\text{alg}}$ framework provides a structured way to reason about liability. If the algorithm replicates prior bias (Match), the source of discrimination may lie in structural or institutional practices. If it amplifies or introduces new disparities (Amplification, Spurious), legal scrutiny may extend to the model's design and deployment. If it reverses or neutralizes disparity (Reversal, Neutralization), this may be viewed either as a corrective mechanism or as a masking of relevant social realities—depending on the normative context.

Overall, this approach enables causal audits that are both technically sound and aligned with legal standards for assessing discriminatory harm.

\paragraph{\textit{Empirical Illustration of  Algorithmic Discrimination Measurement}}

We illustrate the \( \Delta_{\text{alg}} \) framework using the synthetic hiring scenario introduced in~\autoref{sec:mediation}. Recall that \( \Delta_{\text{alg}} \) is defined as the difference between the total causal effect of the sensitive attribute \( A \) on algorithmic predictions \( \hat{Y} \) and on historical human decisions \( Y \):
\[
\Delta_{\text{alg}} = TE_{\hat{Y}} - TE_Y.
\]

We first estimate the total effect of \( A \) on the human hiring outcome \( Y \). Using g-computation, we simulate two counterfactual worlds in which all individuals are assigned \( A=0 \) and \( A=1 \), respectively, while allowing the mediators \( M_1 \) (community service) and \( M_2 \) (residential location) to vary naturally as functions of \( A \). Counterfactual outcomes \( Y(a) \) are generated by propagating mediator values through the structural hiring model and averaged across the population. This yields a small but positive total effect,
\[
TE_Y = 0.0265,
\]
indicating that, after intervening on political affiliation, being liberal is slightly beneficial overall in the historical decision process.

We then compute \( TE_{\hat{Y}} \) for several algorithmic predictors trained on the same data. Each predictor is treated as a black box and excludes \( A \) explicitly, relying instead on different subsets of mediators and covariates. For each algorithm, we simulate counterfactual predictions under \( A=0 \) and \( A=1 \), allowing mediators to respond to the intervention, and compute
\[
TE_{\hat{Y}} = \mathbb{E}[\hat{Y}(1)] - \mathbb{E}[\hat{Y}(0)].
\]

Table~\ref{tab:delta_alg} reports the estimated effects and the corresponding \( \Delta_{\text{alg}} \) categories.

\begin{table}[t]
\centering
\caption{Total effects for human decisions and algorithmic predictions, and resulting \( \Delta_{\text{alg}} \) classifications.}
\label{tab:delta_alg}
\begin{tabular}{lccc}
Outcome & \( TE_Y \) & \( TE_{\hat{Y}} \) & \( \Delta_{\text{alg}} \) Category \\
\midrule
\( \hat{Y}_{\text{match}} \)  & 0.0265 & 0.0265 & Match (Replication) \\
\( \hat{Y}_{\text{proxy}} \)  & 0.0265 & $-0.0842$ & Reversal \\
\( \hat{Y}_{\text{biased}} \) & 0.0265 & $-0.2187$ & Reversal \\
\( \hat{Y}_{\text{fair}} \)   & 0.0265 & 0.0811 & Amplification \\
\bottomrule
\end{tabular}
\end{table}

The matched predictor \( \hat{Y}_{\text{match}} \) closely reproduces the historical effect, yielding \( \Delta_{\text{alg}} \approx 0 \) and indicating that the algorithm merely replicates existing disparities. In contrast, the proxy-heavy predictor \( \hat{Y}_{\text{proxy}} \), which relies primarily on residential location, reverses the sign of the total effect (\( TE_{\hat{Y}} < 0 \)), despite the positive \( TE_Y \). This constitutes a \emph{Reversal} under \( \Delta_{\text{alg}} \), demonstrating how proxy variables can introduce algorithmic harm even when the sensitive attribute is excluded.

The biased predictor \( \hat{Y}_{\text{biased}} \) amplifies this reversal, producing a large negative total effect and representing an extreme case of algorithm-induced discrimination relative to the historical process. Finally, the fairer predictor \( \hat{Y}_{\text{fair}} \), which emphasizes the explaining mediator \( M_1 \), yields a larger positive effect than the human process, corresponding to \emph{Amplification}. Depending on normative and legal context, such amplification may be interpreted either as corrective or as overcompensation.

Overall, these results demonstrate that comparing \( TE_Y \) and \( TE_{\hat{Y}} \) provides a clear and legally interpretable audit of algorithmic behavior. The \( \Delta_{\text{alg}} \) framework isolates whether an algorithm replicates, alters, or introduces discriminatory effects, aligning causal fairness analysis with counterfactual but-for reasoning used in legal assessments of discriminatory harm. The details of the implementation can be found in~\autoref{app:exp_design},~\autoref{app:delta_alg_design}.

\begin{table}[h!]
\centering
\begin{tabular}{|p{3.5cm}|p{5.1cm}|p{5.1cm}|}
\hline
\textbf{Aspect} & \textbf{``\textit{But-for}'' Sensitive Attribute} & \textbf{``\textit{But-for}'' Algorithmic Decision} \\
\hline
\textbf{Core Question} & Would the outcome have been the same if the individual had a different sensitive attribute (e.g., race, gender, address)? & Would the outcome have been the same if the AI system had not been used at all? \\
\hline
\textbf{Causal Object} & Sensitive attribute (e.g., \( A = \) gender, race) & Use of algorithmic system (e.g., \( \hat{Y} = \text{AI decision} \)) \\
\hline
\textbf{Application Context} & Traditional discrimination law (e.g., disparate treatment, redlining) & Algorithmic decision-making and the use of automated systems in high-stakes contexts \\
\hline
\textbf{Evidentiary Basis} & Statistical or causal analysis of how outcomes change with variations in the sensitive attribute & Comparative analysis of human vs. algorithmic decisions; expert evidence or reverse engineering \\
\hline
\textbf{Legal Focus} & Discriminatory treatment based on protected characteristics & Liability for deploying a potentially biased or inaccurate AI system \\
\hline
\textbf{Implication of Positive Result} & Indicates structural or direct discrimination & May suggest algorithm-induced harm or failure to comply with due diligence obligations \\
\hline
\textbf{Causal Tools Used} & Counterfactual inference; total effect of sensitive attribute on outcome & Evaluation of outcome differences under AI use vs. non-use; counterfactual replacement of algorithm \\
\hline
\textbf{Judicial Examples} & \textit{Cook v. HSBC} (credit scoring) & \textit{Pickett}, \textit{Loomis} (risk assessment and sentencing) \\
\hline
\end{tabular}
\caption{Comparison of ``\textit{But-for}'' Sensitive Attribute vs. ``\textit{But-for'}' Algorithmic Decision in Legal Causality}
\label{tab:butfor_comparison}
\end{table}

\section{Accessing Evidence for Causal Auditing}\label{sec:evidence}

Given the practical and theoretical difficulties associated with applying the \textit{but-for} test and counterfactual reasoning in AI-related discrimination cases, a pressing question arises: can evidence disclosure mechanisms support fairer judicial outcomes? More specifically, should litigants have a legally recognized right to access the kinds of evidence necessary for meaningful causal analysis?

From a procedural fairness standpoint, the right to request disclosure is essential for victims of algorithmic discrimination. It provides at least the opportunity to pierce the so-called \textit{opacity veil} - the informational barrier that often obscures the causal chain leading to harm~\citep{grozdanovski2021search}. Recent EU regulatory initiatives suggest a gradual movement toward recognizing this right.

The first major development was the introduction of the \textit{AI Act}~\cite{aiact2024}, a horizontal, cross-sectoral legislative framework that makes two key contributions. First, it introduces a four-tier risk taxonomy for AI systems: non-high, limited, high, and unacceptable. Second, it links these risk levels to specific compliance obligations with technical (safety) standards. For high-risk systems - used in sensitive domains such as employment, education, and finance~\footnote{The eight high-risk sectors listed in Annex III of the AI Act are 1. biometrics (in so far as their use is permitted under relevant Union or national law); 2. critical infrastructure; 3. education and vocational training; 4. employment, workers management and access to self-employment; 5. Access to and enjoyment of essential private services and essential public services; 6. law enforcement (in so far as it is permitted under the relevant Union or national law); 7. Migration, asylum and border control management (when permitted under Union or national law); 8. Administration of justice and democratic processes.} - the Act mandates compliance with standards related to transparency, data governance, and risk mitigation strategies.

Importantly, for high-risk systems, the AI Act includes a \textit{right to an explanation}, enshrined in Article 86 and largely inspired by its counterpart (the right to a human intervention) in Article 22 of the General Data Protection Regulation (GDPR)~\cite{gdpr2016}. 

As argued by Grozdanovski~\cite{grozdanovski2025}  both rights are, without a doubt, laudable steps toward increasing the level of informational transparency in judicial instances that address AI-assisted decision-making. However, for individuals to access the explanations provided under these articles, litigants must meet rather strict requirements, giving the impression that those articles’ application will likely be selective rather than generalized~\cite{grozdanovski2025}. 

In the GDPR’s framework, the "Article-22" explanations are delivered in cases where data subjects have agreed to their data being processed through \textit{fully automated}means, resulting in an output that has legally or significantly affected them. 
The concept of full automation within the meaning of "Article 22" GDPR has been much discussed, the debate in that connection being on whether \textit{fully automated} processing is one performed through automated means and/or if it includes the automatic adherence of the (human) data controller to a specific automated output~\footnote{The so-called Article 29 Working Party (A29WP) seemed to consider both conditions as cumulative, considering that full automation is warranted when a given output is produced through automated means and the ex post human control is practically inexistent (reduced to a ‘token gesture’). See Article 29 Working Party, Guidelines on Automated individual decision-making and Profiling for the purpose of Regulation 2016/679 (3 October 2017, last revised on 6 February 2018), at 21.}. 

More clarity in that connection was expected from the CJEU’s \textit{Schufa case}~\cite{schufa2023}, dealing with an issue of automated credit scoring. On the point of \textit{full automation}, the CJEU considered that that requirement was met in the case at hand, because the processing included the "automated establishment of a probability value based on personal data relating to a person and concerning that person’s ability to repay a loan in the future"~\cite{schufa2023}.  The CJEU did not include, in its assessment, the degree of human reliance on the output, thus failing to specify if the "fulness" of the automated processing should also be examined in relation to the degree of human control and oversight over the features (accuracy, trustworthiness) of an AI’s recommendation. 

In the end, it remains relatively unclear how full automation should be understood, or whether it will be feasible at all for data subjects to adduce evidence of it in cases where they seek access to explanations - for example, regarding (potentially biased) outputs that may have affected the.

Similar observations can be made with regard to the second requirement in Article 22 GDPR which is that a \textit{fully automated} data processing legally or significantly affect a data subject’s personal situation. This requirement is also mentioned in Article 86 AI Act - which sets out its own conditions to access explanations that, on the surface at least, appear to be less stringent than those framing the access to explanations under Article 22 GDPR. 
Indeed, to access Article-86 explanations, proof of ‘full automation’ is not required. Pursuant to Article 86(1) AI act, explanations are due by the deployer - i.e. not the provider - of a high-risk system listed in Annex III, in instances where an output, or lack thereof, has "legally or significantly" affected a person’s health, safety and fundamental rights. 

From the vantage point of knowability, the big question to be raised in relation to both Articles is the following: how can an individual \textit{know} that an AI output has "significantly affected" their fundamental rights? For instance, in the case of automated recruitment, how could a rejected job applicant even suspect that their automated rejection was, in fact, based on an unfair bias (for instance, gender)~\cite{grozdanovski2025}?  

Presumably, it is precisely for the purpose of uncovering the nature and extent of an AI’s effect that that individual would seek an explanation in the first place, and yet, both Articles 22 GDPR and 86 AI Act condition the access to an explanation on the explainees \textit{already} possessing some knowledge (\textit{ergo} evidence) that such an effect had, indeed, been produced.

Moving beyond the procedural requirements that litigants must meet to gain access to explanations under both the GDPR and the AI Act, another relevant question also remains open: \textit{what information} should be included in those explanations? 

With respect to Article 22 GDPR, Wachter et al.~\cite{wachter2017} have seminally argued that the explanations required under this provision are twofold: ex ante - prior to processing, requiring disclosure of information about a system’s functionalities and intended purpose - and ex post - once an output has been generated and encompassing information about the rationale behind the output and the "reasoning" process the system followed in producing it.

The CJEU has shed some light on the substance of the GDPR-explanations in the \textit{Dun}~\cite{dun2025} case where it stated that \textit{meaningful} explanations cover "all relevant information concerning the procedure and principles relating to the use of personal data with a view to obtaining, by automated means, a specific result, the obligation of transparency also requiring that that information be provided in a concise, transparent, intelligible and easily accessible form"~\cite{dun2025}. That information must "describe the procedure and principles actually applied in such a way that the data subject can understand which of his or her personal data have been used in what way in the automated decision-making at issue, with the complexity of the operations to be carried out in the context of automated decision-making not being capable of relieving the controller of the duty to provide an explanation"~\cite{dun2025}.  

Given the current stage of caselaw development in the EU, the \textit{Dun} case can be interpreted as offering a useful \textit{guideline} on the content of explanations that should also be given under Article 86 AI Act, as far as their general relevance and understandability are concerned.

It should, however, be noted that the object of the Article-86 explanations is slightly different than that of Article-22 explanations: the latter (should) uncover the \textit{logic of a given AI system}; the former (should) reveal the rationale of the \textit{human decision to rely (or not) on a specific AI output}. Indeed, Article 86(1) states that deployers of high-risk systems should give explanations that clarify the "\textit{role of the AI system} in the decision-making procedure and the main elements of the decision taken"~\cite{EU_AI_Act_2021}. 

The explanations delivered within the framework of the AI Act do not \textit{prima facie} include the disclosure of technical information about the "inner workings" of an AI system, but a \textit{statement of reasons} aimed at shedding more light on the deployers’ motives to include an AI in a decision-making practice and, eventually, align with its output~\footnote{Explanation pertaining to an AI-assisted human decision seems to be a type of explanation due within the meaning US Blueprint for an AI Bill of Rights, which mentions that a person should know how and why an outcome impacting them was determined by an automated system, “including when the automated system is not the sole input determining the outcome.”  See Blueprint for an AI Bill of Rights. Making Automated Systems Work For The American People (2022), cit. supra, at 40.}.  Because of this, the explanations delivered under Article 86 AI Act appear to be more easily accessible than those delivered under Article 22 GDPR. After all  - and as the \textit{Pickett} case confirms - it is always easier to request from a (human) deployer to explain \textit{their} assessment of a system’s accuracy, fairness and trustworthiness and much more difficult to directly assess those features, when an AI system’s functioning is opaque. 
Be that as it may, CJEU case law on Article 22 GDPR is still emerging, while case law on Article 86 of the AI Act is, at this stage, non-existent. This gives rise to a lingering ambiguity as to how feasible, in practice, the requirements governing access to explanations will be for claimants. It may suffice to stress that, formally at least, the entitlements created by both provisions can be interpreted as a procedural gateway to relevant (causal) knowledge about the features and effects of AI outputs - knowledge that individuals could acquire and deploy when bringing discrimination claims before courts.

What is more, in addition to the right(s) to an explanation, it should be mentioned that AI opacity as barrier to relevant information about AI deployment is at the core of the two EU liability frameworks: the  \textit{AI Liability Directive (AILD)},%
\footnote{Proposal for a Directive of the European Parliament and of the Council on adapting non-contractual civil liability rules to Artificial Intelligence (AI Liability Directive), COM(2022)496 final.} and the Revised Product Liability Directive (PLD)~\cite{pld2024} both of which include a right to request disclosure of evidence. 

It should be noted that the AILD proposal was withdrawn, leaving the amended PLD  as the only of the two instruments to be enacted. Nevertheless, the latter will not be further discussed for a relatively simple reason: the strict liability regime (and system of evidence) it creates does not apply to harm such as discrimination (or any fundamental rights violations). 

The harms the PLD regulates include death or personal injury, medically recognized damage to psychological health~\cite{pld2024};  damage to, or destruction of, any property~\footnote{Art. 6(1)(b), PLD, cit. supra. This provision does not apply in instances where the damaged property is he defective product itself (i); a product damaged by a defective component that is integrated into, or inter-connected with, that product by the manufacturer of that product or within that manufacturer’s control (ii); property used exclusively for professional purposes (iii).} and destruction or corruption of data that are not used for professional purposes~\cite{pld2024}. Had the AILD entered into force, it could have - potentially - applied to human rights violations (including cases of discrimination) as it created a system of civil fault liability, applied in cases of harm occasioned (that is, causally associated) with high-risk systems, within the meaning of the AI Act.

In Article 3, the AILD granted victims a conditional right to request disclosure of evidence. If a defendant - a system provider or deployer - refused to comply, or if a court found the requested evidence to be both relevant and plausible, a rebuttable presumption of liability was triggered. In this way, the AILD attempted to address the information asymmetry between victims and AI system operators.

However, the scope of evidence covered by the AILD was narrower than might be expected. It did not extend to the kinds of technical evidence previously identified as crucial in case law - such as access to system source code, audit logs, or the possibility of reverse-engineering for causal inspection. Instead, the AILD limited evidentiary obligations to demonstrating compliance with the technical requirements set out in the AI Act - for instance, ensuring human oversight, conducting risk assessments, or maintaining documentation.
The underlying logic was that harm - such as discriminatory outcomes - was presumed to arise from non-compliance with the AI Act, implicitly (and ultimately) placing the fault on (human) actors tasked with complying with their duties under that Regulation. Accordingly, legal proceedings are steered away from establishing the full causal chain of algorithmic discrimination and redirected toward identifying failures in human oversight or institutional duty of care. In effect, the AILD narrowed the focus of AI-related litigation: rather than properly and empirically uncovering algorithmic causality, courts would have evaluated whether an operator properly adhered to prescribed standards of responsible AI use.

\section{Practical Constraints and Legal Admissibility of Causal Auditing}\label{sec:constraints}

While our proposed causal auditing framework offers a principled mechanism to quantify algorithmic discrimination, its deployment in real-world settings faces two substantial obstacles: limited data access and judicial skepticism toward technical evidence.

\textbf{Data Access Limitations}
Causal inference techniques, assume access to historical decision data, sensitive attributes, and potential confounders. However, such data are often unavailable due to privacy constraints, proprietary protections, or lack of organizational transparency. To address these barriers, we propose two practical mitigations:

\textbf{Synthetic Data Approximation}: Researchers and auditors can construct synthetic datasets calibrated to known statistical properties of the original data. While such approximations cannot support formal causal identification, they can help estimate plausible bounds of discrimination and generate testable hypotheses for further investigation.

\textbf{External Audits and Intermediaries}: Independent audit frameworks, possibly regulated by courts or supervisory bodies, can be authorized to access proprietary training data and models under confidentiality protections. These auditors can compute causal metrics and produce simplified summaries that are intelligible and admissible in court.

\textbf{Admissibility and Evidentiary Standards}
Legal systems are often cautious toward technically complex methods, especially those perceived as “black boxes.” Courts traditionally require that evidence be:
\begin{itemize}
    \item Understandable to decision makers without technical background  (e.g., judges, juries),
    \item Verifiable by independent experts,
    \item     Admissible under rules such as Daubert in the U.S. or corresponding standards in EU jurisdictions.
\end{itemize}
    
Our proposed $\Delta_{\text{alg}}$ framework is interpretable in principle, but its acceptability in court hinges on the ability to explain how it was derived, what assumptions were made, and what the results mean in lay terms. Courts may prefer simpler statistical associations unless causal models are presented with strong justifications and clarity.

Moreover, judicial skepticism toward algorithmic opacity remains a formidable challenge. As seen in \textit{Pickett and Loomis}, courts were reluctant to demand full transparency or reverse engineering of AI systems due to practicality and cost. Consequently, whether courts will adopt causal metrics such as ours depends on future developments in legal precedent, education, and regulatory scaffolding (e.g., the EU’s AILD procedural presumptions).

To bridge this gap, ongoing dialogue between legal practitioners, AI ethicists, and causal inference experts is essential. Standard-setting bodies and courts may need to adopt minimum standards of explainability and traceability to allow principled causal claims to be judicially actionable.

\section{Conclusions}~\label{sec:conclusion}
\label{sec:conc}

This article has outlined the critical role of causality in evaluating the fairness of AI-driven decisions. We demonstrated that relying exclusively on statistical tools without causal analysis can lead to significant inaccuracies, such as overestimating, underestimating, or even reversing discrimination effects. Furthermore, we highlighted the essential connections between reliable evidence of algorithmic discrimination, legal practices in courts, and evolving regulatory frameworks, particularly within the European Union.

Specifically, we examined how causal reasoning influences legal assessments of discrimination and identified two primary forms of evidence: \textit{direct evidence}, derived from reverse engineering AI systems, and \textit{indirect evidence}. Given that direct evidence is often practically unattainable due to AI opacity, the article explores alternative causal analyses through indirect evidence. Two approaches based on \textit{but-for} counterfactual reasoning were presented:

\textbf{"\textit{\textit{but for}}" Sensitive Attribute} The first approach evaluates the impact of sensitive attributes or redlining mediators in algorithmic decision-making, using test data to investigate causal relationships.

\textbf{"\textit{but for}" Algorithmic Decision} The second approach assesses whether decisions would differ if AI systems were not employed at all, placing emphasis on historical decision-making practices within organizations.

Finally, we introduced a novel causal auditing metric based on comparing the total effects of sensitive attributes on real-world outcomes versus algorithmic predictions and aligned with the "\textit{but for}" algorithmic decision legal framework. This framework provides a legally interpretable metric for identifying bias amplification or attenuation by the algorithm and enables courts to distinguish between structural discrimination and algorithmic liability.

\subsection{Methodological Limitations and Future Research Directions}

While our framework provides a principled basis for identifying and quantifying algorithmic discrimination, its practical application is subject to several methodological limitations.

First, causal inference relies heavily on the accurate specification of causal structures, such as directed acyclic graphs (DAGs). Small changes in assumed causal relationships can lead to large differences in fairness estimates~\cite{plecko2022causal, binkyte2023causal}. The construction of reliable DAGs often requires domain expertise or causal discovery methods, both of which carry uncertainty and may yield inconsistent results~\cite{binkyte2023causal,spirtes2000causation}.

Second, causal analysis in fairness contexts typically depends on observational, rather than experimental, data. This introduces standard challenges such as unmeasured confounding, selection bias, and violations of assumptions like the stable unit treatment value assumption (SUTVA) and ignorability~\cite{rubin2005causal,holland1986statistics}. These limitations complicate efforts to isolate the causal effect of sensitive attributes on outcomes. We outline detailed methodological requirements for applying causal inference for fairness in~\autoref{sec:methodological}.

Third, while our causal auditing framework $\Delta_{\text{alg}}$ can be applied without reverse engineering, it remains an indirect method. It is inherently less probative than direct interpretability approaches, such as mechanistic inspection of model internals. However, recent studies suggest that mechanistic interpretability still faces significant obstacles, including polysemanticity and concept superposition~\cite{lecomteincidental,bereska2024mechanistic}, which limit its current reliability as a source of causal evidence.

Finally, empirical validation of the proposed auditing metric remains an open task. Future work should explore its performance on real and synthetic datasets, assess its robustness to causal misspecification, and evaluate its legal interpretability in applied settings. 

\paragraph{Reproducibility.}
The code for reproducing experiments and figures is available at a public GitHub repository \href{https://github.com/RutaBinkyte/Causal-Case-Against-Algorithmic-Discrimination/tree/main}{Causal-Case-Against-Algorithmic-Discrimination}.

\bibliography{sn-bibliography}

\appendix

\section{When Causality is Needed for Reliably Measuring Discrimination?}~\label{app:causalstructures}

Measuring discrimination without accounting for the underlying causal structure of variable relationships can lead to misleading conclusions and biased estimations of discrimination. In extreme cases, such as Simpson's paradox, this bias can even reverse conclusions—for example, a biased estimation may suggest positive discrimination when, in reality, an unbiased analysis would reveal negative discrimination.

\begin{figure}[H]
\begin{minipage}{0.32\linewidth}
\centering
		{\includegraphics[scale=0.6]{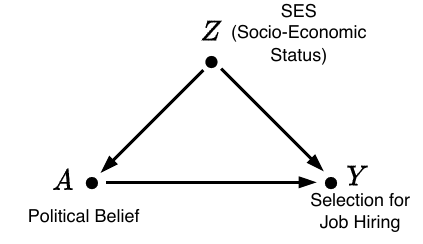}}
	\caption{Confounder structure.}
	\label{fig:simpleConfJobHiring1}
\end{minipage}
\hspace{0.3cm}
\begin{minipage}{0.3\linewidth}
\centering
  \includegraphics[scale=0.6]{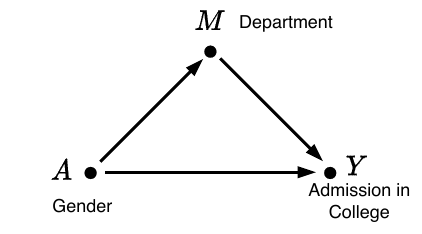}
	\caption{Mediator structure.}
	\label{fig:simpleMediatorAdmission1}
 \end{minipage}
 \hspace{0.3cm}
\begin{minipage}{0.3\linewidth}
\centering
  \includegraphics[scale=0.6]{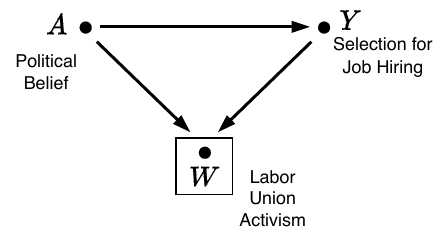}
	\caption{Collider structure.}
		\label{fig:simpleColliderJobHiring1}
\end{minipage}
\end{figure}

Figures~\ref{fig:simpleConfJobHiring1}-\ref{fig:simpleColliderJobHiring1} illustrate the three fundamental causal structures that can result in statistical anomalies, making common fairness metrics unreliable.

\noindent

\subsection{Confounder structure}
\label{sec:confExample}

The first scenario in which ignoring the causal structure of the data can lead to an unreliable estimation of discrimination arises from failing to account for a confounding variable. Consider the hypothetical example in Figure~\ref{fig:simpleConfJobHiring1}, which depicts an automated system used to select candidates for job positions.

Suppose the system takes two input features: socio-economic status (SES), denoted as $Z$, and the candidate’s political belief, denoted as $A$. The outcome, $Y$, represents whether the candidate is selected for the next stage of hiring (or the probability of selection). The SES influences the hiring outcome because candidates from higher SES backgrounds may have greater access to reputable academic institutions and expensive training programs.

Both SES ($Z$) and political belief ($A$) can be either binary (e.g., rich vs. poor for SES, liberal vs. conservative for political belief) or continuous (e.g., the degree of wealth for SES and the extent of conservatism for political belief). In this example, we assume that a candidate's political belief is influenced solely by their SES.

Now, suppose the automated decision system is suspected of being biased against candidates with a particular political belief. The claim is that the system is more likely to select candidates who hold a specific political ideology. However, failing to account for the confounding effect of SES may lead to incorrect conclusions about whether the system is truly biased.

A simple approach to assessing the fairness of the automated selection process, represented by $Y$, with respect to the sensitive attribute $A$, is to compare the conditional probabilities:  

\[
\pr(Y=1 \mid A=0) \quad \text{and} \quad \pr(Y=1 \mid A=1)
\]

This comparison, known as statistical disparity, quantifies the difference in selection rates between the two groups (e.g., conservatives and liberals). However, such an estimation of discrimination is biased due to the confounding effect of $Z$.  

Since $Z$ influences both the sensitive variable $A$ and the outcome $Y$, it introduces a correlation between $A$ and $Y$ that is not causal. In other words, candidates with a high socio-economic status (SES) are more likely to have conservative political beliefs and, at the same time, have a higher probability of being selected due to their access to better academic institutions and training opportunities. This results in an observed correlation in the data: employers will see a greater proportion of candidates with conservative political beliefs and a lower proportion of those with liberal beliefs. However, this correlation is driven by the confounder $Z$ and should not be misinterpreted as discrimination.  

Most statistical fairness metrics, such as equal opportunity and predictive parity, fail to account for such confounding effects, making them unsuitable for measuring discrimination in the presence of these statistical anomalies.

\subsection{Mediator structure}
\label{sec:medExample}

The second scenario in which failing to account for the causal structure of data leads to unreliable discrimination estimation arises from the presence of one or more mediator variables. The core issue is whether discrimination transmitted through a mediator variable should be considered justifiable or not. Similar to confounding structures, a mediator variable can result in Simpson's paradox.  

A well-known example of Simpson's paradox caused by a mediator structure is the gender bias observed in the 1973 Berkeley graduate admissions study~\citep{berkeley75,loftus18}. Figure~\ref{fig:simpleMediatorAdmission} illustrates the causal graph underlying the data, where the sensitive variable ($A$) represents gender, the outcome ($Y$) represents admission to Berkeley graduate programs, and a single mediator variable ($M$) represents the department to which a candidate applied.  

In 1973, 44\% of male applicants were admitted, compared to only 34\% of female applicants. At first glance, this appears to indicate a bias against female candidates. However, when the data was analyzed separately by department, acceptance rates were found to be approximately equal across genders.  

In a simple mediator structure, there are two possible causal paths from $A$ to $Y$: a direct path, $A\rightarrow Y$, and an indirect path, $A\rightarrow M\rightarrow Y$. Comparing the overall admission rates of male and female candidates considers both paths in the discrimination measure. In contrast, comparing admission rates within each department isolates only the direct path, $A\rightarrow Y$. Consequently, whether or not to account for mediator paths when measuring discrimination can lead to contradictory conclusions, as demonstrated by Simpson's paradox.

\subsection{Collider structure}
\label{sec:collExample}

A biased estimation of discrimination can also arise due to the presence of a common effect (collider) variable and implicit conditioning on that variable during the data generation process. Using the same hypothetical job selection example, consider the causal graph in Figure~\ref{fig:simpleColliderJobHiring}, where $A$ and $Y$ retain the same meanings as in the previous example.  

Assume that the data used to train the automated decision system is collected from various sources, but primarily from labor union records. Let $W$ represent a candidate’s labor union activism, which is influenced by both $A$ and $Y$. On one hand, political belief ($A$) affects whether a candidate becomes an active labor union member, as individuals with liberal political beliefs are more likely to join labor unions. On the other hand, if a candidate is selected (hired), they are more likely to become a labor union member, increasing the likelihood that their case appears in labor union records. Following prior work, a box around a variable ($W$) indicates that data is implicitly conditioned on that variable during the data collection process.  

Once again, the simple approach of comparing selection rates between the two groups (conservatives and liberals) leads to a biased estimation of discrimination due to the colliding path through $W$. Intuitively, an individual appears in the collected data either because they hold liberal political beliefs or because they were selected for the job. Candidates who are both liberal and selected remain in the data, but conditioning on labor union activism introduces a non-causal correlation between $A$ and $Y$. Specifically, the dataset drawn from labor union records includes fewer liberal candidates who were selected for the job compared to conservative candidates. This pattern may misleadingly suggest discrimination against liberal candidates. However, this correlation arises from the collider structure and should not be mistaken for actual discrimination.

\section{Experimental Design and Synthetic Data Details}
\label{app:exp_design}

\subsection{Causal Graph and Variables}
\label{app:graph}

Our experimental setup follows the mediation structure in Figure~4 (main text), adapted from Binkytė et al.~\cite{binkyte2022need}. The sensitive attribute is \(A\in\{0,1\}\) denoting political belief (\(0=\) conservative, \(1=\) liberal). The outcome \(Y\in\{0,1\}\) denotes a hiring decision. We include a confounder \(C\) representing socio-economic status (SES) (conceptually \(0=\) high SES, \(1=\) low SES), and two mediators: \(M_1\) (community service participation/hours) and \(M_2\in\{0,1\}\) (address, \(0=\) rural, \(1=\) urban). The graph encodes four pathways between \(A\) and \(Y\):
\[
A \leftarrow C \rightarrow Y,\qquad
A \rightarrow Y,\qquad
A \rightarrow M_1 \rightarrow Y,\qquad
A \rightarrow M_2 \rightarrow Y.
\]
The confounding path \(A \leftarrow C \rightarrow Y\) is non-causal with respect to interventions on \(A\), whereas the remaining paths are causal.

\subsection{Data Generation Process}
\label{app:dgp}

We simulate \(n\) individuals from a structural causal model consistent with the graph. The data-generating process is designed to satisfy three properties: (i) raw hiring rates exhibit negative statistical parity against \(A=1\), (ii) there exists direct discrimination against \(A=1\) (negative direct effect), and (iii) there are offsetting positive mediated effects through \(M_1\) (and to a smaller extent \(M_2\)), so that the total causal effect can differ from raw disparity.

We generate data as follows:
\begin{align}
A &\sim \text{Bernoulli}(0.5), \nonumber \\
C &\sim \mathcal{N}(0,1), \label{eq:dgp_base}\\
M_1 \mid A,C &= \alpha_0 + \alpha_A A + \alpha_C C + \varepsilon_{1}, \quad \varepsilon_{1}\sim\mathcal{N}(0,\sigma_1^2), \label{eq:dgp_m1}\\
M_2 \mid A,C &\sim \text{Bernoulli}\!\left(\sigma(\beta_0 + \beta_A A + \beta_C C)\right), \label{eq:dgp_m2}\\
Y \mid A,C,M_1,M_2 &\sim \text{Bernoulli}\!\left(\sigma(\gamma_0 + \gamma_A A + \gamma_C C + \gamma_1 M_1 + \gamma_2 M_2)\right), \label{eq:dgp_y}
\end{align}
where \(\sigma(\cdot)\) is the logistic sigmoid. Parameter signs are chosen to reflect the narrative: (i) \(A\) is associated with SES via correlation through \(C\) (capturing \(A\leftarrow C\)), (ii) \(\gamma_A<0\) encodes direct bias against \(A=1\), (iii) \(\alpha_A>0\) and \(\gamma_1>0\) encode a positive indirect pathway \(A\rightarrow M_1\rightarrow Y\), and (iv) \(\beta_A>0\) and \(\gamma_2>0\) provide a smaller positive pathway via \(M_2\).

\paragraph{Observed disparity (statistical parity).}
We report the raw (associational) group disparity:
\[
SP = \Pr(Y=1\mid A=1) - \Pr(Y=1\mid A=0).
\]
In the generated dataset used in the main experiments, we obtain \(\Pr(Y=1\mid A=1)=0.289\), \(\Pr(Y=1\mid A=0)=0.373\), hence \(SP=-0.084\), i.e., liberals are hired less often in observed data.

\subsection{Mediation Analysis Estimation Procedure}
\label{app:mediation_est}

To estimate mediation quantities in Section~\ref{sec:mediation} (main text), we use a parametric imputation approach under sequential ignorability. We fit the following models on the synthetic data:
\begin{align}
\widehat{M}_1 &= \widehat{\mathbb{E}}[M_1\mid A,C] \quad\text{(OLS)}, \nonumber\\
\widehat{M}_2 &= \widehat{\Pr}(M_2=1\mid A,C) \quad\text{(logistic)}, \nonumber\\
\widehat{Y} &= \widehat{\Pr}(Y=1\mid A,C,M_1,M_2) \quad\text{(logistic)}. \label{eq:mediation_models}
\end{align}
We then impute counterfactual mediator values \(\widehat{M}_j(a)\) by setting \(A=a\) while keeping \(C\) fixed at its observed value, and propagate these through the outcome model to obtain counterfactual outcome probabilities. Using population averages, we compute:
\begin{align*}
TE &= \mathbb{E}[Y(1)] - \mathbb{E}[Y(0)],\\
NDE &= \mathbb{E}[Y(1, M_1(0), M_2(0))] - \mathbb{E}[Y(0, M_1(0), M_2(0))],\\
NIE_{M_1} &= \mathbb{E}[Y(0, M_1(1), M_2(0))] - \mathbb{E}[Y(0, M_1(0), M_2(0))],\\
NIE_{M_2} &= \mathbb{E}[Y(0, M_1(0), M_2(1))] - \mathbb{E}[Y(0, M_1(0), M_2(0))].
\end{align*}
For the synthetic dataset reported in the main text, the estimates are:
\[
TE \approx 0.0272,\quad
NDE \approx -0.1173,\quad
NIE_{M_1} \approx 0.1349,\quad
NIE_{M_2} \approx 0.0271,
\]
illustrating that a negative direct effect is offset by a large positive mediated effect via \(M_1\).

\subsection{Experimental Design Delta Y}
\label{app:delta_alg_design}

To audit the algorithm’s contribution to disparity, we treat the human decision \(Y\) and algorithmic prediction \(\hat{Y}\) as distinct endpoints and compare their total causal effects:
\begin{align}
TE_Y &= \Pr(Y=1\mid do(A=1)) - \Pr(Y=1\mid do(A=0)), \label{eq:te_y}\\
TE_{\hat{Y}} &= \Pr(\hat{Y}=1\mid do(A=1)) - \Pr(\hat{Y}=1\mid do(A=0)). \label{eq:te_yhat}
\end{align}
We then compute \(\Delta_{\text{alg}} = TE_{\hat{Y}} - TE_Y\) and map the pair \((TE_Y,TE_{\hat{Y}})\) to a taxonomy category (Match, Amplification, Mitigation, Reversal, Neutralization, Spurious, Baseline).

\paragraph{G-computation / Monte Carlo estimator.}
We estimate \(\Pr(\cdot\mid do(A=a))\) via Monte Carlo g-computation. We fit mediator models \(M_1\mid A,C\) and \(M_2\mid A,C\) (as in \autoref{eq:mediation_models}). Under the intervention \(do(A=a)\), we keep \(C\) fixed at its empirical distribution and sample mediators:
\[
M_1^{(s)} \sim \mathcal{N}(\widehat{\mathbb{E}}[M_1\mid A=a,C], \widehat{\sigma}_1^2), \qquad
M_2^{(s)} \sim \text{Bernoulli}(\widehat{\Pr}(M_2=1\mid A=a,C)),
\]
for \(s=1,\dots,S\) Monte Carlo draws. We then plug \((A=a,C,M_1^{(s)},M_2^{(s)})\) into the relevant endpoint model (human or algorithm) and average predicted probabilities over draws and individuals to approximate \(\Pr(\cdot\mid do(A=a))\). In our implementation we use \(S=80\) draws per individual.

\subsection{Prediction Algorithms}
\label{app:pred_algorithms}

All predictors are treated as black boxes and are evaluated solely through the counterfactual effects in \autoref{eq:te_yhat}. Predictions are produced by logistic regression decision rules with a \(0.5\) threshold. Crucially, the predictors differ by feature choice and/or the strength of a penalty on \(A\), enabling distinct \(\Delta_{\text{alg}}\) behaviors.

\begin{itemize}
    \item \textbf{Human process model (\(Y\), audit baseline).}
    We fit a logistic regression approximation of the historical decision:
    \[
    \Pr(Y=1\mid A,C,M_1,M_2) = \sigma(\theta_0 + \theta_A A + \theta_C C + \theta_1 M_1 + \theta_2 M_2).
    \]
    This model is used as the endpoint for \(TE_Y\) estimation.

    \item \textbf{Matched predictor (\(\hat{Y}_{\text{match}}\)).}
    A predictor designed to replicate the human model’s functional form and inputs:
    \[
    \Pr(\hat{Y}_{\text{match}}=1\mid A,C,M_1,M_2) = \sigma(\theta_0 + \theta_A A + \theta_C C + \theta_1 M_1 + \theta_2 M_2).
    \]
    This serves as a near-replication benchmark.

    \item \textbf{Proxy-heavy predictor (\(\hat{Y}_{\text{proxy}}\)).}
    A predictor emphasizing the proxy \(M_2\) and including \(A\):
    \[
    \Pr(\hat{Y}_{\text{proxy}}=1\mid A,M_2) = \sigma(\phi_0 + \phi_A A + \phi_2 M_2).
    \]
    This model operationalizes proxy-driven behavior that can diverge sharply from \(Y\) under intervention.

    \item \textbf{Biased predictor (\(\hat{Y}_{\text{biased}}\)).}
    A deliberately biased variant obtained by adding an extra negative penalty to the proxy model’s linear predictor:
    \[
    \Pr(\hat{Y}_{\text{biased}}=1\mid A,M_2) = \sigma(\phi_0 + \phi_A A + \phi_2 M_2 + \lambda A),
    \]
    with \(\lambda<0\) (implemented as an additional penalty on \(A\)).

    \item \textbf{Fairer predictor (\(\hat{Y}_{\text{fair}}\)).}
    A predictor that explicitly excludes \(A\) and the proxy \(M_2\), relying on \(C\) and the explaining mediator \(M_1\):
    \[
    \Pr(\hat{Y}_{\text{fair}}=1\mid C,M_1) = \sigma(\psi_0 + \psi_C C + \psi_1 M_1).
    \]
\end{itemize}

\section{Methodological Constraints in Using Causality for Fairness and Possible Solutions}
\label{sec:methodological}

In this section, we outline key methodological requirements for applying causal inference, particularly those most relevant to fairness applications. In addition, we discuss how they can be satisfied in practice.

\subsection{Availability of Causal Graph}

The effective use of causal inference in fairness applications is shaped by several critical methodological requirements and assumptions. Among these, one of the most significant is the availability of an accurate \textit{directed acyclic graph} (DAG) representing relationships among variables. \footnote{The DAG is subject to additional assumptions, including the \textit{causal Markov condition}, \textit{causal faithfulness}, and \textit{causal sufficiency}. Together, these conditions encode the same requirements as \textit{SUTVA} and \textit{ignorability} in the potential outcome framework and, therefore, will not be discussed separately.}~\citep{pearl2009causality}. Research by~\citep{binkyte2023causal} highlights significant disagreements in causal fairness estimations due to minor variations in the assumed causal structure.  

The availability of a DAG is particularly crucial in the presence of \textit{collider structures}, as conditioning on a collider induces bias in causal effect estimation~\citep{pearl1988probabilistic}. Additionally, DAGs are essential for evaluating \textit{path-specific effects}, which play a key role in distinguishing between \textit{redlining} and \textit{explaining variables} in fairness scenarios.  

Causal structures (or causal graphs) can be obtained through two main approaches: expert consultation or learning from observational data. However, both approaches have limitations. Domain experts may hold differing views or introduce biased assumptions.  

Learning causal relationships from data, on the other hand, often requires additional assumptions regarding data distribution, functional dependencies, relationships between exogenous unobserved variables, and an informed choice of the learning algorithm. Research by~\citep{binkyte2023causal} demonstrates how different causal discovery algorithms, when applied to the same dataset, can yield significantly different causal structures. This variability raises concerns about the reliability of causality learned purely from observational data~\citep{guyon2011causality}.  

Expert-driven causal graphs, while useful, are not easily scalable and may be prone to human error. However, recent advancements in \textit{causal discovery with large language models (LLMs)} offer a promising alternative~\citep{kiciman2023causal, kasetty2024evaluating, vashishtha2023causal}. Hybrid approaches that combine LLM-driven insights with statistical causal discovery methods appear to be particularly promising in bridging the gap between expert-driven and data-driven causal inference~\citep{afonja2024llm4grn}.

\subsection{Causal Transportability}
Beyond the accurate identification of causal graphs, another crucial consideration when applying causal inference to fairness is whether causal findings obtained in one context remain valid and generalizable across diverse populations—an issue addressed by the concept of \textit{causal transportability}. Currently, the application of causal knowledge across populations relies heavily on assumptions about general causal structures that remain relatively stable across different contexts. Examples include physical laws, such as "altitude causes temperature," or well-established medical principles, such as "bacteria cause disease."  

However, the principles underlying ethical AI are often deeply intertwined with cultural values and societal norms, which are neither immutable nor universally consistent. These variations present significant challenges in transferring causal insights across different demographic or geographic populations.  

The framework developed by Pearl and Bareinboim utilizes \textit{directed acyclic graphs} (DAGs) to adjust causal knowledge for new settings through targeted data collection~\citep{bareinboim2014transportability, pearl2011transportability}. Building on this, \citet{binkyte2024babe} propose an \textit{expectation-maximization} (EM) approach to adapt causal knowledge for applications in target demographic groups, further enhancing the practicality of causal transportability in real-world scenarios.

\subsection{Possibility for Intervention}

In fairness estimation, the sensitive attribute is typically considered the exposure or treatment variable, with the goal of measuring its impact on the outcome.  

Most definitions of causal effect are based on the notion of intervention or manipulation of a cause variable (exposure)~\citep{vanderweele2012causal,holland1986statistics}. This presents a challenge when making causal claims about non-manipulable attributes, such as race or gender.  

Some approaches in the literature suggest shifting the focus from actual manipulation to changes in perception~\citep{rahmattalabi2022promises}. For instance, instead of altering the gender of candidates to estimate its effect on hiring decisions, researchers could manipulate the employer’s \textit{perception} of gender. This could be achieved by submitting two otherwise identical résumés with different names or titles indicating different genders. Such an approach aligns with methodologies used in social experiments examining the impact of race or gender on hiring decisions~\citep{neumark2018experimental}.  

\citep{vanderweele2012causal} further differentiates immutable sensitive attributes into those that are randomized at birth (e.g., biological sex) and those that are not (e.g., race, social gender). This distinction is important when estimating the causal effect of a sensitive attribute. If the attribute is randomized, its causal effect on an outcome can be estimated by directly comparing exposure levels. For example, the total causal effect of biological sex on an outcome can be estimated by comparing observed differences between men and women~\citep{vanderweele2012causal}.  

In contrast, race is not randomly assigned but influenced by ancestral and socio-cultural factors, making causal effect estimation more complex. Even at the biological level, estimating the effect of race requires a thorough understanding of the causal structure of relevant covariates. Such estimation is particularly relevant in medical scenarios, where assuming independence between the sensitive attribute and the outcome (e.g., disease probability) is unreasonable.  

Similarly, in potential discrimination scenarios, \citep{vanderweele2012causal} and \citep{rahmattalabi2022promises} suggest shifting the focus to the direct effect of the \textit{perceived} gender or race on decision-making. This perspective aligns with causal fairness approaches that assess discrimination based on how sensitive attributes are perceived and acted upon within decision-making processes.

\subsection{Causal assumptions}
 The \textit{Stable Unit Treatment Value Assumption} (SUTVA)~\citep{rubin1986statistics} imposes two key requirements: \textit{no interference} and \textit{consistency}.  

The \textit{no interference} assumption states that the effect of the sensitive attribute on the outcome should not be influenced by interactions between individuals. However, in social sciences research, where interactions and feedback loops are common, this assumption is often challenged, necessitating careful discussion and restricted interpretations of causal estimates~\citep{morgan2015counterfactuals}.  

Since fairness is typically evaluated within a social context, potential interactions must be carefully assessed. For example, in the hiring scenario, a violation of the \textit{no interference} assumption would occur if hiring individuals from one political spectrum increased the likelihood of favoring candidates with the same political beliefs in future hiring decisions. This scenario is plausible because current employees may prefer candidates who share their political ideology, creating a feedback loop that perpetuates bias.  

The \textit{consistency} assumption requires that each treatment level leads to the same potential outcomes~\citep{keele2015statistics}. In fairness evaluation, the treatment variable is replaced by the sensitive attribute, which is often a socially constructed category such as race or gender. Identifying the causal effect of gender on hiring becomes problematic if gender itself does not have a consistent effect on hiring decisions. For instance, discrimination may not apply uniformly to all women; rather, only women who exhibit a certain level of "femininity" might experience bias. If this possibility cannot be ruled out, it should be considered in studies aiming for a fine-grained causal analysis.  

In summary, SUTVA assumptions are likely to be violated in fairness scenarios. However, causal approaches can still be applied if the results are interpreted with caution. Some methods for identifying causal effects despite SUTVA violations are discussed in~\citep{laffers2020identification}.

 
 
The \textit{ignorability} assumption~\citep{rubin1986statistics} requires that the sensitive attribute and the outcome be independent, given the observable variables. In other words, no unobserved variables should create a significant link between the sensitive attribute and the outcome.  

In fairness evaluation, the presence of such a link could indicate that a portion of the observed discrimination is, in fact, a spurious effect induced by a confounding variable. For example, if education is a confounder but is not included in the dataset, its confounding effect cannot be controlled. Consequently, it becomes impossible to estimate the causal effect of political belief on hiring decisions separately from the effect of education.  

Unobserved confounders are less likely to exist for immutable sensitive attributes such as sex or race, as these attributes generally do not have temporally prior causes. However, dependency between noise terms related to the sensitive attribute and the outcome can still violate ignorability.  

\citep{fawkes2022selection} highlight the implications of assuming ignorability when using causal counterfactuals. Following their reasoning, consider the case of college admissions (illustrated in Figure~\ref{fig:simpleMediatorAdmission}). Under the ignorability assumption, an average male applicant to a technical profession could be counterfactually replaced with an average female applicant to the same profession. However, due to social expectations tied to gender roles, a woman applying to a technical profession is likely to be more motivated and hardworking than an average male applicant with the same professional aspirations. This example illustrates how violations of ignorability can introduce bias into causal estimations.

The \textit{positivity} assumption~\citep{rubin1986statistics} is violated if certain combinations of a sensitive attribute and a covariate have zero probability. Violations of positivity can be either \textit{deterministic} or \textit{random}~\citep{westreich2010invited}.  

For example, positivity would be violated if a specific level of education always corresponded to liberal political beliefs. In this scenario, the violation would likely be \textit{random}, as it is improbable that education would have a strictly deterministic relationship with political beliefs. In such cases, statistical methods exist for handling analysis under violations of positivity~\citep{westreich2010invited}.  

However, consider a case where obtaining a Harvard degree is treated as an explanatory mediator between ethnicity and hiring. Due to historical patterns of discrimination and socioeconomic disparities, certain ethnic groups may have \textit{near zero probability} of obtaining a Harvard degree. This constitutes a \textit{deterministic} violation of positivity. In such cases, it is essential to reconsider whether possessing a Harvard degree should be an absolute requirement for the job in question, given its potential for exclusion and disparate impact.


The identifiability of path-specific effects in the presence of multiple mediators requires the absence of causal links between the mediators~\citep{vanderweele2014mediation}. Evaluating path-specific effects is particularly important for understanding the mechanism through which the sensitive attribute influences the outcome. As discussed earlier (Section~\ref{sec:medAnalysis}), the effect of the sensitive attribute can be considered either justifiable or discriminatory depending on the mediating variables along the causal path.  

However, in fairness scenarios, causal dependencies between mediators are common. For example, consider a case where the relationship between race and hiring decisions is mediated by both social status (e.g., redlining) and education (an explaining variable). It is highly probable that education level is influenced by social status. In such a scenario, the indirect effects of race through social status and education separately are not identifiable, as the mediators are causally dependent.  

To address this challenge,~\citep{vanderweele2014mediation} proposed a method that treats multiple mediators jointly. In some cases, this approach can help disentangle and identify individual indirect effects in the presence of causal links between the mediators.

Recognizing these methodological challenges underscores the importance of carefully balancing technical rigor and practical applicability. With these considerations in mind, we now summarize our findings and highlight future directions for research.

\end{document}